%% file: tmlr.tex
\documentclass[10pt]{article} 
\usepackage[preprint]{tmlr}

\input{math_commands.tex}

\usepackage{hyperref}
\usepackage{url}
\usepackage{amsthm}

\usepackage{float}
\usepackage{natbib}
\usepackage{tabularx}
\usepackage{algorithm}
\usepackage{algorithmic} 
\usepackage{url}
\usepackage{booktabs}
\usepackage{xcolor}
\usepackage{subcaption}

\usepackage{amssymb}
\usepackage{tikz}
\usepackage{multirow}
\usepackage{amsmath}
\usepackage{makecell}

\usepackage{microtype}
\usepackage{graphicx}
\usepackage{subcaption}
\usepackage{booktabs} 
\usepackage[utf8]{inputenc}
\usepackage[T1]{fontenc}

\usepackage{hyperref}
\usepackage{amsmath}
\usepackage{amssymb}
\usepackage{mathtools}

\usepackage[capitalize,noabbrev]{cleveref}
\usepackage[capitalize,noabbrev]{cleveref}
\usepackage{amsthm}
\newtheorem{observation}{Observation}
\usepackage[textsize=tiny]{todonotes}
\usepackage{graphicx}

\title{Position: Tensorization is a powerful but
underexplored tool for compression and
interpretability of neural networks}

\author{\name Safa Hamreras \email safa.hamreras@dipc.org \\
      \addr Donostia International Physics Center, Paseo Manuel de Lardizabal 4, E-20018 San
Sebastián, Spain
      \AND
      \name Sukhbinder Singh \email sukhi.singh@multiversecomputing.com \\
      \addr Multiverse Computing, Spadina Ave., Toronto, ON M5T 2C2, Canada
      \AND
      \name Roman Orus \email roman.orus@multiversecomputing.com\\
      \addr  Multiverse Computing, Paseo de Miramón 170, E-20014 San Sebastián, Spain \\ Ikerbasque Foundation for Science, Maria Diaz de Haro 3, E-48013 Bilbao, Spain \\ Donostia International Physics Center, Paseo Manuel de Lardizabal 4, E-20018 San
Sebastián, Spain}

\def\month{MM}  
\def\year{YYYY} 
\def\openreview{\url{https://openreview.net/forum?id=XXXX}} 

\begin{document}

\maketitle

\begin{abstract}
Tensorizing a neural network corresponds to reshaping some or all of its dense weight matrices
into higher-order tensors and approximating them
using \emph{low-rank tensor network} decompositions.
This technique has shown promise as a model
compression strategy for large-scale neural networks. However, despite encouraging empirical
results, tensorized neural networks (TNNs) remain underutilized in mainstream deep learning.

In this position paper, we offer a perspective on
both the potential and current limitations of TNNs.
We argue that TNNs represent a powerful yet underexplored framework for deep learning—one that deserves greater attention from both engineering and theoretical communities. Beyond compression, we emphasize TNNs as a flexible class of architectures with distinctive scaling behavior and additional structural tools for interpretability. We introduce a
new conceptual tool—the \emph{stack representation}—
which shows that any single tensor network layer
is equivalent to a sparse, structured deep linear
network. A defining feature of TNNs is the presence of \emph{bond indices}, which create latent spaces
absent in conventional networks. These internal
representations offer a new lens for tracking how
features evolve across layers, with potential implications for mechanistic interpretability. We conclude by outlining key research directions aimed
at addressing the practical challenges of scaling
and deploying TNNs in modern deep learning
systems.

\end{abstract}
\section{Introduction}

\begin{figure}[t]
    \centering
    \includegraphics[width=0.6\columnwidth]{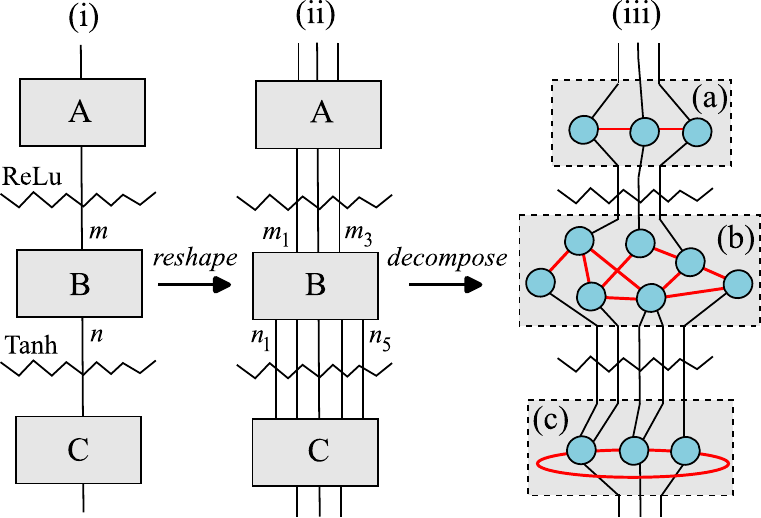}
    \caption{
    (i) An MLP with three fully-connected layers $A, B, C$, depicted as rectangles, and two non-linear layers (here, ReLU and Tanh activation layers), depicted as wiggly lines. The network is tensorized as follows.
    (ii) First, each weight matrix ($A, B$ and $C$) is reshaped into a higher-dimensional tensor. For example, the $m \times n$ matrix $B$ can be reshaped into an 8-index tensor with 3 input indices $m_1, m_2, m_3$ such that $m_1 m_2 m_3 = m$ and 5 output indices $n_1, n_2, n_3, n_4, n_5$ such that $n_1 n_2 n_3 n_4 n_5 = n$.
    (iii) Then each such tensor is decomposed as a tensor network, i.e., a contraction ($\texttt{einsum}()$) of tensors. Three examples of tensor network layers are shown: (a) a matrix product operator (or tensor train) layer, (b) a generic tensor network layer and (c) a tensor ring layer. A tensor network decomposition exposes new degrees of freedom inside the neural network, carried by the bond indices of the network, highlighted in red.
    }
    \label{fig:tensorization}
\end{figure}

If neural network scaling laws hold, achieving greater intelligence will require continued increases in model size—and cost \cite{kaplan2020scaling}. However, such growth cannot
continue indefinitely. Two main strategies address this challenge. The first is to compress large, already-trained dense
models without significantly degrading performance \cite{zhu2024survey}, an approach for which several effective methods exist. The second, more fundamental strategy is to design neural networks with more efficient parameterizations. In fact, the success of high-quality compression methods
indicates that modern neural networks are highly overparameterized. This raises a central question: which degrees of
freedom actually carry information or high-level knowledge
in dense networks? Identifying these degrees of freedom
could enable the design of more compact and efficient network parameterizations.

Large language models are now routinely compressed using highly effective methods. Their weights can often be pruned—with little or no loss in performance—including, in
some cases, entire layers \cite{sun2023simple, ma2023llm, sanh2019distilbert}. Compact, high-performing models can
also be obtained through quantization \cite{xiao2023smoothquant, liu2024llm} or by distilling smaller models from larger
ones \cite{hinton2015distilling}.

Beyond these standard approaches, dense weight matrices
in neural networks often admit low-rank tensor network
(TN) decompositions along non-obvious directions. This
enables the replacement of some or all dense layers with
structured TN layers, see Fig. \ref{fig:tensorization}. Architectures that combine
dense and tensorized components are known as tensorized
neural networks (TNNs). As a compression technique, tensorization is complementary to more traditional methods
such as quantization \cite{xiao2023smoothquant, liu2024llm},
pruning \cite{sun2023simple, ma2023llm}, and knowledge
distillation \cite{hinton2015distilling}. Crucially, tensorization can
be combined with these other techniques to achieve higher
overall compression \cite{polino2018model, aghli2021combining, zeng2024novel}.

The effectiveness of these compression strategies suggests
that only a small subset of layers or even individual neurons
strongly influences the model’s output \cite{maini2023can, yu2024super, ghorbani2020neuron}. Arguably, training
a large model and applying compression afterward may be
the most effective strategy, likely reflecting limitations of
current gradient-based optimization methods.

In this position paper, \textbf{we argue in Sec. \ref{position} that TNNs merit
greater attention not only as a compression technique,
but as a distinct and promising class of deep neural
networks.}

First, when effective for compression, tensor network decompositions reduce parameter counts while simultaneously
revealing fine-grained correlation structure in the weights,
encoded in the geometry of the tensor network.\footnote{Tensor networks originated in quantum physics as compact
representations of many-body wavefunctions with structured correlations \cite{white1992density, orus2019tensor, cirac2021matrix}.}

The tensor network structure of a model may emerge from
pretrained weights---varying across layers and architectures—or be hardwired as an inductive bias during training. In the latter case, an appropriate TN architecture can
improve both training efficiency and accuracy. This expectation is motivated by the fact that real-world data is highly
structured and correlated \cite{bermudez2018analysing, lu2025tensor, chen20252, dkebowski2015relaxed, lin2016criticality, solgi2025tensorized}. As a result, the activations of a well-trained network should retain these correlations,\footnote{This is evident in CNNs, where intermediate activations often
resemble structured transformations of the input.} implying that the learned weights must preserve the data’s intrinsic structure.

Second, although information flow between layers remains
sequential, the internal tensor network structure within each
layer can induce complex, non-sequential computation patterns. This also leads to a much richer architectural and
hyperparameter space than standard dense layers, providing
greater design flexibility.

To support a clearer analysis of TNNs, we introduce a new
conceptual tool—the \emph{stack representation}—which interprets
their inductive bias as structured sparsity in the weights. We
also discuss how TNNs naturally enable acceleration of
both forward and backward passes during training and inference. In addition, we highlight their potential for advancing
mechanistic interpretability, building directly on the stack
representation.

In Sec. \ref{alternative}, we explain why TNNs, despite these advantages,
have not yet seen widespread adoption. We identify the
main obstacles to their development and outline research
directions to address them. Finally, we present a vision for
fully \emph{tensorized neural networks}, in which all components—
including weights, data, activations, nonlinearities, and normalization layers—are represented using tensor networks.
We argue that such architectures could underpin the next
generation of scalable and efficient neural networks.

This article is timely because tensorized neural networks
(TNNs) address three converging challenges in deep learning: scaling, theory, and interpretability. As model sizes
grow to unprecedented scales and conventional scaling laws
show signs of saturation, TNNs offer a well-developed alternative framework with qualitatively different scaling behavior. Their controllable \emph{bond dimensions} recover dense
models in a limit, providing mathematically grounded parameters for probing capacity and generalization. At the
same time, TNNs embed decades of advances and insights
about tensor networks originating from quantum many-body
physics directly into neural architectures. Techniques originally developed to understand entanglement, renormalization, and hierarchical representations in physical systems
are directly relevant for explaining and analyzing deep models. (See discussion around Fig. \ref{fig:forward_tensorization} (ii).) With tensor network
machine learning already demonstrating the effectiveness
of tensor networks as stand-alone architectures \cite{stoudenmire2016supervised, rieser2023tensor}, now is the right
moment for both the machine learning and tensor network
communities to adopt TNNs as both practical and theoretical
drivers of the next stage in deep learning.

The remainder of the paper is organized as follows. In Sec. \ref{background}, we give a technical background of TNNs. 
In Sec. \ref{position},
we present our case for why tensorized neural networks
are a compelling class of models that merit further study.
In Sec. \ref{alternative}, we discuss the opposing view, focusing on the limitations that currently hinder large-scale deployment of TNNs.

\section{Technical background review of tensorized neural networks}\label{background}

We review tensor network-based layers for neural networks by focusing mostly on the \textit{Matrix
Product Operator}(MPO) linear layer and tensorized convolutional layers based on \textit{Tucker} and \textit{Canonical Polyadic}
(CP) decompositions. These will serve as our main examples, however, our discussion extends to other TN variants.
Many alternative decompositions have been studied, each
with different trade-offs. For a broader survey, see \cite{kolda2009tensor} for tensor decomposition and its applications
that may extend to neural networks, and \cite{wang2023tensor}
for neural network tensorization.

\begin{figure}[h]
    \centering
    \includegraphics[width=\textwidth]{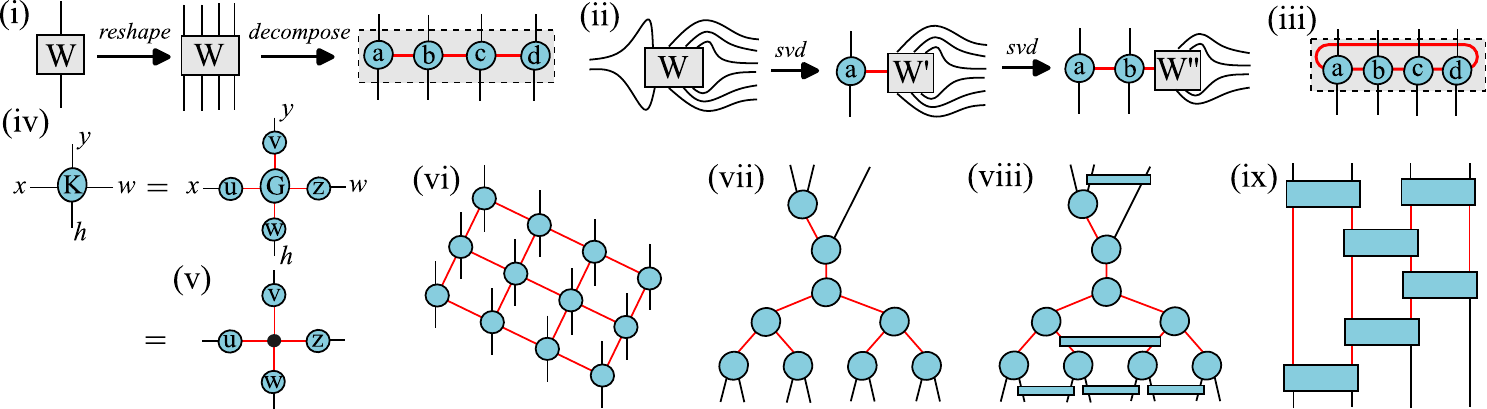}
    \caption{
    Examples of tensor network layers.
    (i--ii) An MPO (or tensor train) layer, obtained by reshaping a pretrained weight matrix $W$ into a higher-order tensor and decomposing it via repeated singular value decompositions. The special case with all bond dimensions equal to 1 (no red line) corresponds to a hypercube of neurons that are completely uncorrelated across different hypercube dimensions.
    (iii) A Tensor Ring layer, which can capture more correlations than an MPO with the same bond dimension.
    (iv) A Tucker decomposition of a 2D convolution kernel $K_{xywh}$, where indices $x, y$ and $w, h$ denote the spatial location of the top-left corner of an image patch and the patch’s width and height, respectively; the bond dimensions correspond to the Tucker ranks.
    (v) A Canonical Polyadic decomposition of $K$, a special case of Tucker where the core tensor $G$ reduces to a delta tensor.
    (vi--ix) Quantum physics–inspired layers: PEPS, Tree TN (hierarchical Tucker decomposition), a Quantum Circuit (comprised of orthogonal gates), and a MERA layer (which combines a tree structure with orthogonal gates).
    }
    \label{fig:tn_examples}
\end{figure}

\subsection{Tensorizing fully-connected linear layers using Matrix Product Operators}
\textit{A Matrix Product Operator} (MPO), also known as a \textit{Tensor Train} \cite{oseledets2010tt}, is a TN decomposition of a matrix as a set of tensors arranged on
a line. Each tensor connects to its immediate neighbors,
except for the first and last tensors, which connect only
to the second and second-last tensors, respectively. For
example, Fig. \ref{fig:tn_examples}(i) shows the decomposition of a dense
weight $W$ as an MPO with four tensors a,b,c, and d
as $W_{i_1 \dots j_4} = \sum_{\chi \chi' \chi''} a_{i_1 j_1 \chi} \, b_{\chi i_2 j_2 \chi'} \, c_{\chi' i_3 j_3 \chi''} \, d_{\chi'' i_4 j_4}$. In practice, this decomposition is obtained by iteratively applying SVDs as described in App.~A; see Fig. \ref{fig:tn_examples}(ii). Note that the SVD ranks $|\chi|$, $|\chi'|$, $|\chi''|$---the bond dimensions of the MPO---are not directly related to the standard SVD rank of the matrix $W$; a full-rank matrix in the usual sense can have small MPO bond dimensions, while a low-rank matrix might have large ones. Instead, the MPO bond dimensions reflect the strength of ``spatial'' correlations between the weights once the weight matrix is arranged in a higher-dimensional hypercube---in this case, an 8-dimensional array $W_{i_1 \dots j_4}$. More generally, an MPO decomposition of the weight matrix $W$ reshaped to have $n$ input and $n$ output indices is $W = \sum_{\chi} W^{[1]}_{i_1 j_1 \chi_1} \left( \prod_{n=2}^{N-1} W^{[n]}_{\chi_{n-1} i_n j_n \chi_n} \right) W^{[N]}_{\chi_{N-1} i_N j_N}$.
Recently, MPO layers have been used to compress large
language models, see, e.g., \cite{tomut2024compactifai, xu2023tensorgpt, yuan2023asvd}.

\subsection{Tensorizing Convolution Layers}

A commonly used TN decomposition for compressing convolution layers is the Tucker decomposition
that was introduced in \cite{tucker1966some} and widely applied in literature \cite{kim2015compression, phan2020stable, hayashi2019exploring}. Here, the 4-index convolution kernel $K_{xywh}$ is decomposed into a 4-index \emph{core tensor} $G$ and four \emph{factor matrices} $u, v, w, z$ as $K_{xywh} = \sum_{\chi_x \chi_y \chi_w \chi_h} G_{\chi_x \chi_y \chi_w \chi_h} \, u_{x,\chi_x} \, v_{y,\chi_y} \, w_{w,\chi_w} \, z_{h,\chi_h}$, see Fig. \ref{fig:tn_examples}(iv). The dimensions of the internal (red) indices are the \emph{Tucker ranks}. Additionally, convolution kernel compression can also be achieved through the \emph{Canonical Polyadic (CP)} decomposition \citep{harshman1970foundations,carroll1970analysis}, as in \citep{lebedev2014speeding,astrid2017cp,zhou2019tensor}. This approach expresses the convolution kernel $K$ as a sum of rank-1 tensors as $K = \sum_{r=1}^{R} \lambda_r \left( a_r^{(x)} \otimes a_r^{(y)} \otimes a_r^{(w)} \otimes a_r^{(h)} \right)$ where $R$ is the CP rank of the tensor $W$. This decomposition can be seen
as a special case of the Tucker decomposition, where the
core tensor is a delta tensor, Fig. \ref{fig:tn_examples}(v). In practice, convolution layers decomposed using Tucker or CP formats
often perform well even with relatively low ranks \cite{gabor2023compressing, zhou2019tensor}, including on large-scale, real-world tasks \cite{martin2023boosting}. See
also the review in \cite{wang2023tensor}. Notably, pre-trained
convolution kernels frequently exhibit low-rank structure
under Tucker and CP decompositions \cite{singh2024tensor}.
Convolutional kernels can also be factorized based on SVD
decomposition \cite{jaderberg2014speeding}, where 2D convolutions are replaced with two smaller 1D convolutions, and
network decoupling \cite{guo2018network}, by transforming regular convolutions into depth-wise separable convolutions
speeding up the network.

Finally, tensorizing convolutional layers can also help mitigate CNNs’ limited generalization and vulnerability to noise
or adversarial attacks by introducing regularization techniques—such as tensor dropout—that have been proposed
to enhance robustness while preserving the multilinear structure \cite{kolbeinsson2021tensor, chen2024machine}.

\subsection{Other notable TN layers}

Several other types of tensor network (TN) layers have
been explored. Some prominent examples, which we do
not focus on here, are the tensor ring layer \cite{zhao2016tensor}, used to develop tensor ring nets \cite{wang2018wide},
tensor train \cite{white1993density}, and tensor train cross approximation \cite{oseledets2010tt}. Adaptive TN
geometries have also been studied \cite{hashemizadeh2020adaptive, solgi2024low}. Inspired by quantum many-body systems, we highlight several other TN layer architectures,
illustrated in Fig. \ref{fig:tn_examples}(vi)-(ix), that merit exploration: (1) Projected Entangled Pair Operator (PEPO) layers \cite{verstraete2004renormalization, lin2021low, o2020simplified},
a higher-dimensional generalization of MPOs; (2) Tree Tensor Network operators \cite{shi2006classical, murg2010simulating, liu2019machine, cheng2019tree, milbradt2024state},
also known as the Hierarchical Tucker Decomposition \cite{oseledets2009breaking, grasedyck2010hierarchical}; (3) Quantum Circuit layers composed of orthogonal gates; and (4) the
Multi-scale Entanglement Renormalization Ansatz (MERA)
\cite{vidal2008class}, which extends tree TNs by incorporating
additional disentangling tensors. These physics-inspired
layers are grounded in tensor networks that have yielded
powerful insights into the structure of quantum many-body
wavefunctions, suggesting promising inductive biases for
neural networks. For example, tree and MERA architectures
have deep connections to the renormalization group, offering a scale-resolved representation of wavefunctions. (As
discussed later, this scale resolution may help enhance the
interpretability of sparse autoencoders.) These layers have
already proven effective as stand-alone models. Finally,
quantum circuit layers open the door to hybrid quantum-classical neural networks, where bottleneck layers in classical models could be accelerated by executing them on
quantum devices \cite{aizpurua2024quantum}.

\section{Position: TNNs are awesome}\label{position}

In this section, we present our main argument. Tensorized
neural networks form a broad class of structured architectures. To keep the discussion concrete, we focus primarily
on TNNs built from matrix product operator (also known
as tensor train) layers. However, the arguments apply more
generally to other TNN variants.

\textbf{(1) Tensorization injects a relevant inductive bias into
neural networks.}

An inductive bias sets a hypothesis about training data that
can be generalized to unseen data \cite{utgoff2012machine, goyal2022inductive}. A general framework for understanding
inductive biases is provided by \emph{geometric deep learning}
\cite{masci2016geometric}, which, for instance, connects many
inductive biases to enforcing symmetries on the weights.
For instance, the success of convolutional neural networks
can be traced to the fact that their layers are designed to
respect translation equivariance \cite{mei2021learning}. The inductive bias introduced by TNNs appears to have a different
origin, not one rooted in obvious symmetries. Instead, it can
be understood as resulting from constraints on the “spatial”
correlations between the weights.\footnote{As mentioned previously, weights inside a tensorized layer
are spatially arranged in a high-dimensional hypercube (this corresponds
to reshaping the weight matrix as a higher-order tensor).
The bond dimensions of a TN then inform the correlations between
different directions of the cube. In general, a higher bond dimension
allows for more complex correlations between weights, while
a smaller bond dimension restricts them. Bond dimension equal to
one, Fig. \ref{fig:tn_examples}(iii), corresponds to the absence of correlations.}

But why should we expect the weights to be correlated
in this way? As previously mentioned, real-world data is
often highly structured and exhibits strong correlations. It
is therefore reasonable to expect that the learned weights of
a neural network also reflect this structure – suggesting that
relatively small bond dimensions in the corresponding tensor
network may suffice. This idea also aligns with the view
that gradient descent tends to align the statistical structure
of the data, weights, and gradients throughout training (see,
e.g., \cite{radhakrishnan2024mechanism}). The main challenge for
tensorization is then to identify tensor networks that best
reflect the correlation structure of the weights in a given
layer.

\textbf{(2) The “Stack representation” of tensorized layers.}

Interpreting inductive bias in terms of weight correlations
may seem abstract. Below we develop a more intuitive—
and interpretability-relevant—“stack representation" of tensorized layers, according to which a tensor network layer is a
sparse and structured \emph{deep linear network} \cite{saxe2013exact, li2021statistical}. This representation can be used
to analyze the inductive bias and other properties of TNNs.

For instance, consider the MPO layer shown in Fig. \ref{fig:tn_examples}(i).
We now describe how the MPO bond indices are not just
internal wiring that encode structured correlations between
neurons, but serve as meaningful feature-carrying channels
that propagate representations through the network; they are
semantically active components in the computation.

\begin{figure}[t]
    \centering
    \includegraphics[width=0.9\columnwidth]{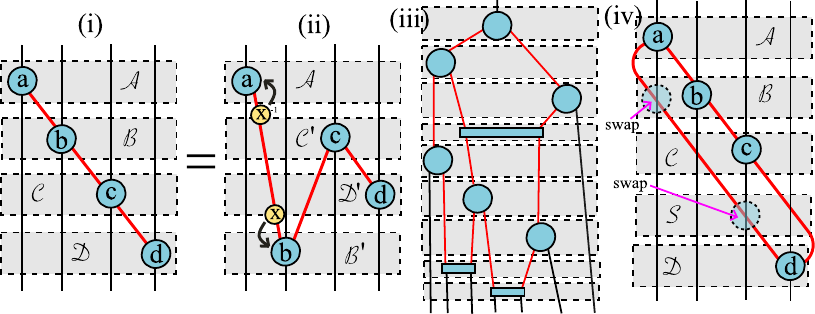}
    \caption{
    Stack representations of TN layers.
    (i) A stack representation of an MPO layer consisting of 4 tensors as a composition of 4 sparse fully-connected layers $\mathcal{A}, \mathcal{B}, \mathcal{C},$ and $\mathcal{D}$. The bond indices of the MPO simply become the input and output dimensions of standard linear layers.
    (ii) Another distinct stack representation of the same MPO layer.
    (iii) A stack representation of a MERA-like layer.
    (iv) A stack representation of a Tensor Ring layer. The ring structure introduces swap operations (layers $\mathcal{B}, \mathcal{S}$) inside the stack.
    }
    \label{fig:stack_representation}
\end{figure}

Fig. \ref{fig:stack_representation}(i) illustrates how the MPO layer can be interpreted as a sequence or stack of standard---but sparse---fully-connected layers $\mathcal{A}, \mathcal{B}, \mathcal{C}$. Each of these sparse layers is the tensor product of a matricization of an MPO tensor and identity matrices, e.g., $\mathcal{B} = I^{(\mathrm{out})}_1 \otimes \hat{b} \otimes I^{(\mathrm{in})}_3 \otimes I^{(\mathrm{in})}_4$, where $\hat{b}$ is a matricized version of the MPO tensor $b$, and $I^{(\mathrm{out})}_1$, $I^{(\mathrm{in})}_3$, $I^{(\mathrm{in})}_4$ are identity matrices with dimensions matching the first output, third input, and the fourth input indices of the MPO, respectively.

An MPO can be represented by multiple, distinct yet equivalent stacks of fully-connected layers. One way to construct an alternative stack is by shifting the order of the MPO tensors in the stack, as illustrated in Fig. \ref{fig:stack_representation}(ii). This produces linear layers based on different matricizations of the same set of MPO tensors. For example, we may obtain a layer of the form $\mathcal{B}' = I^{(\mathrm{out})}_1 \otimes \hat{b}' \otimes I^{(\mathrm{out})}_3 \otimes I^{(\mathrm{out})}_4$, where $\hat{b}'$ is an alternative matricization of tensor $b$. Another source of flexibility comes from transforming to different but equivalent MPO tensors by inserting a factorization of identity, $X X^{-1} = I$, along one or more bond indices of the MPO (potentially using a different factorization for each bond). The matrices $X$ and $X^{-1}$ can then be absorbed into the neighboring MPO tensors prior to building the stack by matricizing the tensors. Transforming the MPO tensors in this
way does not change the weight matrix that results from
contracting the MPO.\footnote{In the quantum physics literature, this representational freedom inherent in an MPO is referred to as the MPO gauge freedom
(Orús, 2019; Cirac et al., 2021).}

The two forms of representational flexibility described
above may have important implications for the interpretability of TNNs—a topic we revisit later in this section. Similar
analyses of inductive bias can be carried out for other tensor
network architectures, see Fig. \ref{fig:stack_representation}(iii–iv). In fact, the stack
representation is geometry-agnostic and extends naturally to
diverse tensor network geometries. However, the presence
of loops—such as in a Tensor Ring—introduces additional
swap operations in the stack as illustrated in Fig. \ref{fig:stack_representation}(iv). The
presence of these swaps operations underscores the qualitatively different inductive bias that a Tensor Ring layer
imparts to the network when compared against an MPO
layer. Note also that the stack view of the MERA does not
contain any swap operations despite the presence of loops.
This is because, unlike the MERA, the Tensor Ring is not a
planar tensor network; projecting it on the plane introduces
crossings that can be removed by continuous deformations.
More broadly, the geometry of the underlying tensor network strongly shapes the nature of the inductive bias. This
makes TNNs a rich design space, offering a spectrum of
biases to tailor for different tasks. With careful selection,
an appropriate tensor network structure has the potential to
meaningfully accelerate learning.

\textbf{(3) Tensorization for expressive yet compact models.}

\begin{observation}[Expressivity and compression trade-off]
Let a dense weight matrix be tensorized into an MPO with $K$ tensors and bond dimension $\chi$. In this case, $\chi$ controls the trade-off between compression and expressivity, where smaller $\chi$ values reduce the expressivity of the MPO but makes it more efficient, and larger  $\chi$ values enhance the expressive power of the MPO at the cost of having more parameters. When $\chi$ is small enough, then the MPO has fewer parameters than the original weight matrix, while possibly achieving comparable accuracy, demonstrating an effective trade-off between compression and expressivity.
\end{observation}

Consider an MPO layer composed of three tensors 
$a_{i_1 j_1 \chi}$, $b_{\chi i_2 j_2 \chi'}$, and $c_{\chi' i_3 j_3}$. 
The total number of parameters is given by
\[
N_{\text{MPO}} = i_1 j_1 \chi + \chi i_2 j_2 \chi' + \chi' i_3 j_3.
\]
The contraction of the MPO gives a dense weight matrix with input indices $(i_1, i_2, i_3)$ and output indices $(j_1, j_2, j_3)$, with
\[
N_{\text{Dense}} = i_1 j_1 i_2 j_2 i_3 j_3.
\]
When the compression ratio $N_{\text{MPO}} / N_{\text{Dense}} \ll 1$, the MPO provides a substantially more compact representation of the original layer.

\noindent\textbf{Example.} 
Consider a weight matrix of size $1000 \times 1000$, reshaped as 
$(10 \times 10 \times 10) \times (10 \times 10 \times 10)$, and decomposed into three MPO cores. 
Assuming a uniform bond dimension $\chi = 20$, the number of parameters becomes
\begin{align*}
N_{\text{MPO}} &= (10 \cdot 10 \cdot \chi) 
               + (\chi \cdot 10 \cdot 10 \cdot \chi) 
               + (\chi \cdot 10 \cdot 10) \\
               &= 2000 + 40000 + 2000 \\
               &= 44000.
\end{align*}
In contrast, the dense representation requires
\[
N_{\text{Dense}} = 1000 \times 1000 = 10^6
\]
This corresponds to a reduction of $95.6\%$, illustrating the strong compression achieved by the MPO structure. The bond dimension $\chi$  controls this trade-off: smaller values yield higher compression.

\cite{gao2020compressing} in Fig. 2 demonstrate that the MPO bond dimension controls the compression--expressivity trade-off on MNIST dataset. As the bond dimension increases, the test accuracy improves and approaches that of the dense model. Similarly,  many prior works have proven that tensorization leads to a better parameter efficiency while maintaining a good expressivity for several classes of tensor networks and neural networks, reflected by accuracy ( see \cite{wang2023tensor}). For instance, in TensorGPT \cite{xu2023tensorgpt}, Fig. 3(a–d) shows that substantial parameter reduction can be achieved with negligible loss in accuracy and related metrics, indicating that tensorization effectively exploits redundancy in model parameters. Moreover, larger models exhibit greater robustness to compression, suggesting that tensorization becomes increasingly effective at scale.  Other similar experiments demonstrating parameter efficiency have been reported in prior work. For instance, Fig. 2 in \cite{lebedev2014speeding} illustrates the application of CP decomposition to convolutional neural networks (CNNs), while Tables I–IV in \cite{liu2023efficient} present results on tensorized transformer architectures, both maintaining good accuracy.

\begin{figure*}[t]
    \centering
    \includegraphics[width=\textwidth]{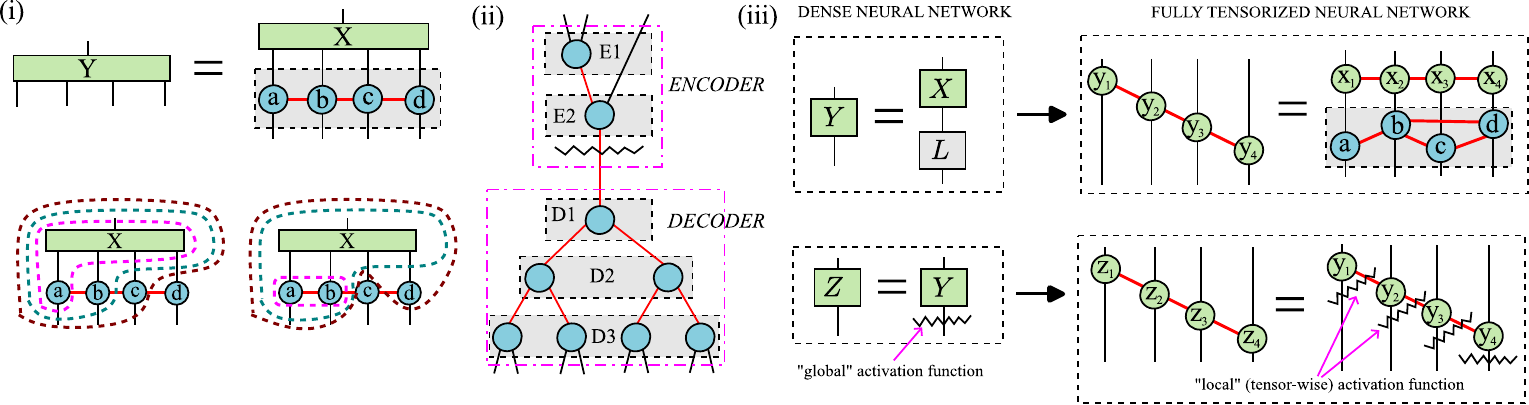}
    \caption{
    (i) Forward pass of a dense activation $X$ through an MPO layer involves a multi-tensor $\texttt{einsum}$ resulting in the activation $Y$. Depicted also are two different contraction orders to perform the $\texttt{einsum}$ as indicated by the two sequences of nested contours.
    (ii) The stack view of a tensorized Sparse Auto-Encoder.
    (iii) Dense vs Fully Tensorized Neural Network: (Top) Forward pass of an activation $X$ through a dense layer $L$ producing a dense activation $Y = LX$ corresponds to forward pass of a tensorized activation through a tensorized layer producing a tensorized activation. (Bottom) A usual non-linear function $f$ (wiggly line) acts component-wise on a dense activation $Y$ resulting in activation $Z = f(Y)$, while a fully tensorized neural network consists of ``local'' activation functions that act tensor-wise, producing tensorized activations.
    }
    \label{fig:forward_tensorization}
\end{figure*}

We provide a simple experiment to illustrate that TNNs are parameter-efficient models that preserve the model’s performance.

Specifically, we compare a standard convolutional neural network (CNN) with its tensorized counterpart. 

\paragraph{Baseline CNN.}
As a baseline, we consider a simple CNN architecture composed of:
\begin{itemize}
    \item a convolutional layer ($1 \rightarrow 32$ channels, $3 \times 3$ kernel),
    \item a ReLU activation,
    \item a max-pooling layer (downsampling by a factor of 2),
    \item a fully connected layer for classification with 10 output classes, the weight matrix has the shape $(6272,10)$.
\end{itemize}

This model serves as the reference base architecture. Please note that this architecture is not optimized for compactness. Our goal is instead to provide a simple  baseline to validate our claims. In particular, we aim to show that tensorized neural networks (TNNs) can significantly reduce the number of parameters while maintaining competitive performance.

\paragraph{Tensorized Neural Network (TNN).}
The tensorized model follows the same high-level structure, but replaces standard layers with tensorized counterparts:

\begin{itemize}
    \item \textbf{Tucker convolution layer:} The convolutional kernel is decomposed using the Tucker format into three components:
    \begin{itemize}
        \item an input projection implemented as a $1 \times 1$ convolution,
        \item a core convolution with a $3 \times 3$ kernel,
        \item an output projection implemented as a $1 \times 1$ convolution.
        \item Tucker ranks are $(1,8)$.
    \end{itemize}

    \item \textbf{MPO layer:} The final classification layer is represented as a two-core MPO. The input shapes are $(112, 56)$ and the output shapes are $(2, 5)$ and the bond dimension is set to 32.
\end{itemize}

The ranks of tucker layer and MPO are chosen to achieve a compression rate of 69.86\%, which represents the relative reduction of parameters.

\paragraph{Training setup.}
Both models are trained on the MNIST training set for 5 epochs using the cross-entropy loss. We use the Adam optimizer with a learning rate of $10^{-3}$ and a batch size of 128. Experiments are conducted on Google Colab Pro+ \footnote{\url{https://colab.research.google.com}} with GPU acceleration (A100). They are evaluated on the MNIST test set of 10,000 samples.

\paragraph{Results and discussion.}

\begin{figure}[H]
    \centering
    \includegraphics[width=0.55\columnwidth]{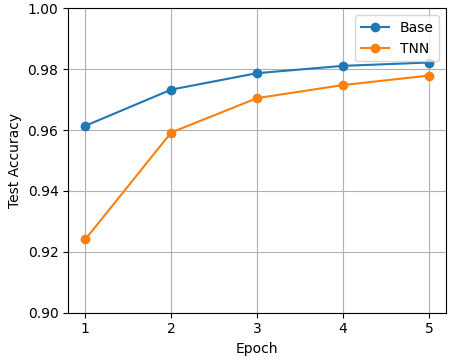}
    \caption{Test Accuracy Across Training Epochs: Base vs. TNN. }
    \label{fig:mnist}
\end{figure}

Fig. \ref{fig:mnist} illustrates the evolution of test accuracy during training for both base and tensorized models. The base model converges faster, and achieves a higher accuracy early in training, the best accuracy is achieved at epoch 5, where it equals 0.9822 using 63,050 parameters.  The TNN improves significantly after the first epoch and achieves a competitive test accuracy at epoch 5 compared to the base model, which equals 0.9779 using 19,002 parameters. This means a reduction in parameters of 69.86\% with less than 0.5\% accuracy degradation in favor of the TNN. 

Overall, these results demonstrate that tensorization provides an effective mechanism for reducing model complexity while preserving most of the expressive power, supporting its suitability for building compact yet accurate neural networks.

\textbf{(4) TNNs have a richer hyperparameter space and can
be scaled in several directions.}

A standard dense fully-connected layer can be scaled by simply enlarging the dimensions (number of hidden neurons) of
the weight matrix. In a convolutional layer, scaling involves
increasing the number of channels in the convolution kernel.
An attention layer offers two different scaling directions –
either expanding the dimensions of the weights matrices
inside the heads and / or increasing the number of attention
heads. On the other hand, TNNs offer more flexible scaling
directions.

\begin{observation}[Scaling Behavior of Tensorized Neural Networks]
Tensorized Neural Networks (TNNs) exhibit favorable parameter scaling relative to their dense counterparts. Furthermore, scaling can be performed along multiple independent axes---bond dimension, tensor order, and network geometry---providing a richer scaling regime than standard neural networks.
\end{observation}

Empirical evidence shows that TNNs achieve comparable performance with fewer parameters, suggesting favorable scaling laws compared to standard neural networks.
TNNs exhibit subtle interactions between depth,
width, and bond dimension, which can be clarified by means
of the stack representation according to which tensorized
layers can be understood as deep linear networks.

Firstly, the width of an MPO layer (that is, the number of
input and output neurons of the layer) can be increased in
two ways—(1) either by enlarging the dimensions of the
existing MPO tensors or (2) by inserting new tensors while
keeping the original tensors intact. Remarkably, the first
approach effectively increases both the width and the depth
of the network, since each additional tensor contributes an
extra factor in the stack. \footnote{Of course, this increase is not strictly proportional to the number of added tensors, since MPO layers lack nonlinearities.} The second approach increases
width in a more conventional sense, without adding depth.
The approach (1) may also have practical applications. In
the context of incremental training \cite{gepperth2016incremental}, one can freeze a pretrained TNN and train only the
newly added tensors on new data. This strategy may help
mitigate catastrophic forgetting—a central challenge in incremental learning.

Bond dimensions play a different role: while they appear
as internal indices of the layer, they more directly map to
width. For example, in Fig. \ref{fig:stack_representation}(i), the output dimension of
$\mathcal{A}$ in the stack is \emph{larger} than its input dimension whenever
the leftmost bond dimension exceeds one. Thus, the bond
dimension primarily controls the effective width not only of
the stack but also of the whole network.

While bond dimensions govern the effective \emph{width} of tensorized architectures, depth provides an additional and complementary source of expressivity, i.e. increasing the depth of a TNN while keeping its
width fixed enhances expressivity in a manner analogous to
conventional neural networks. Each tensorized layer applies
a distinct transformation to the hidden state, enabling hierarchical feature extraction across layers—often with fewer
parameters per layer than in dense architectures. Each added
tensorized layer offers finer choices—choice of tensor network geometry, number of tensors, and the bond dimensions.
This added flexibility enables more precise architectural tuning but also increases the design complexity. The enhanced
flexibility in scaling TNNs may also lead to novel scaling
laws—which strongly depend on the model’s architecture
and inductive bias \cite{tay2023scaling}—and also possibly new
emergent abilities \cite{wei2022emergent}.

\textbf{(5) Tensorization offers a novel model compression approach, which can be seamlessly combined with traditional compression strategies.}

While tensorization is an effective compression strategy in
its own right \cite{hawkins2021bayesian, gabor2023compressing, ma2022lightweight}, one of its key advantages is its compatibility with other standard compression methods – such
as pruning, quantization, and distillation \cite{hinton2015distilling}
– making it a complementary and flexible approach. When
combined, these techniques can yield significantly higher
compression rates. For example, tensorization can be directly applied after pruning or quantization, or a TNN can be
distilled from a dense, possibly already compressed, model. However,
systematically investigating the optimal ways to integrate
tensor networks with other compression strategies remains
an open and important research direction (see Sec. \ref{alternative} for
further discussion).

\textbf{(6) Forward pass acceleration.}

The forward pass through an MPO layer proceeds by reshaping the input matrix X and contracting it with the MPO,
as illustrated in Fig. \ref{fig:forward_tensorization}.\footnote{Depending on how the contraction is carried out, the output tensor $Y$ may need to be reshaped into a matrix before it is
returned.} A naive way to carry out this
computation is to first contract all the MPO tensors to reconstruct the full dense weight matrix M, following a pairwise
contraction sequence such as $M = (((ab)c)d)$, and then compute $Y = MX$. However, this method will always be slower than the standard forward pass through a dense fully-connected layer with the same input and output dimensions as the MPO as it essentially recreates the dense layer before applying it. Alternatively, we can perform the MPO contraction in different pairwise sequences, e.g., $Y = ((((Xa)b)c)d)$ or $Y = (((X(ab))d)c)$, see Fig. \ref{fig:forward_tensorization}(i),
not all of which are equally expensive. The cost of each
sequence depends heavily on the dimensions
of the MPO indices; some sequences can be significantly
cheaper than others. 

It is important to note that tensorization does not always yield acceleration in practice, but rather that it enables — at least in principle — acceleration in regimes where the tensor structure and other factors allow it. Tools like \textit{opt-einsum}\footnote{\url{https://pypi.org/project/opt-einsum/}} can be used to find efficient contraction paths automatically. However, actual runtime performance also depends on factors like hardware (CPU vs. GPU) and low-level optimizations \cite{liu2024high}. For example, a contraction that’s fast using hand-written matrix multiplications may run slower via a general-purpose \textit{einsum()} call. Some contraction sequences can also be executed directly as a sum of products without forming intermediate matrices. In the future, specialized hardware such as FPGAs or ASICs could be leveraged to accelerate these custom contraction patterns (see, e.g., \cite{liu2021efficient}).
While more research is needed to improve current forward pass strategies for TNNs and to design new, more efficient ones, in principle, TNNs offer the potential for substantial speedups in the forward pass \cite{lebedev2014speeding}.

\textbf{(7) TNNs offer additional, distinctive tools for interpretability.}

A central goal in deep learning is to understand how neural networks work and to develop interpretative narratives explaining how they produce their outputs. While significant progress has been made in this direction \citep{fan2021interpretability}, a general and comprehensive understanding remains out of reach. We next argue that TNNs provide additional structural tools for mechanistic interpretability that are not available in standard dense architectures—not because their internal features are inherently more class-selective than a dense layer's, as we show below, but because the tensor network decomposition exposes internal structure (multiple bond spaces, equivalent alternative stacks) that a dense layer simply does not have.

\textit{Interpreting TNNs.}

The unconventional latent spaces associated with bond indices in a TNN offer new avenues for interpretability. As
described in Sec. \ref{position} and illustrated in Fig. \ref{fig:stack_representation}, these bond
indices correspond to the input and output dimensions of
the internal linear layers in the “stack” view of a TNN. As
information propagates through the network, these bond
spaces – like traditional feature spaces – are expected to
develop task-relevant representations. These intermediate
bond features can thus be viewed as a “temporal” resolution
of the total features output by the MPO layer. Owing to
the fact that the same MPO layer can be decomposed into
multiple equivalent stacks of linear layers, different stacks
correspond to different temporal resolutions of the same
output, offering a fine-grained perspective on the network’s
internal representations.\footnote{When two stacks are related simply by gauge transformations
– i.e., insertions of identity resolutions as shown in Fig. \ref{fig:stack_representation}(iii) –
the corresponding intermediate bond features are connected via
multiplication by a matrix. When stacks arise from different matricizations of the same MPO tensor, the resulting features are related
by more general linear maps that effectively bend tensor indices
(“dualize” vectors to forms and vice-versa).} Studying how these bond features relate to output features may provide additional tools for analyzing intermediate representations in tensorized neural networks.

Consider a tensorized layer (e.g., an MPO) applied to a given input data. Using the stack representation, this layer can be viewed as a sequence of sparse linear transformations. This enables the analysis of intermediate features in a manner analogous to neurons. Below, we provide two experiments for bond features analysis. These experiments are
conducted on Google Colab Pro+ with GPU acceleration (A100). They are evaluated on the MNIST test set
of 10,000 samples.

\textit{Experiment 1: Class Selectivity of bond features.}

To investigate whether individual bond dimensions show class-selective activation patterns, we trained a simple tensorized neural network consisting of two linear layers, an MPO layer, and a final classification layer. The detailed architecture is given below: 

\begin{itemize}
    \item \textbf{Input layer:} Flattened MNIST images with 784 input features (28x28).

    \item \textbf{Linear layer 1:} 784 input features and 256 output features, followed by a ReLU activation.

    \item \textbf{Linear layer 2:} 256 input features and 128 output features, followed by a ReLU activation.

    \item \textbf{Two-core MPO layer:}
    \begin{itemize}
        \item The first core has an input index of 16, and an output index of 4.
        \item The second core has an input index of 8, and an output index of 4.
        \item The bond dimension $\chi=32$, which is the maximum bond dimension for this MPO architecture, allowing higher feature selectivity.
    \end{itemize}

    \item \textbf{Linear layer 3 (classifier):} 16 input features and 10 output features, which equals the 10 MNIST digit classes.

\end{itemize}

The model was trained for 20 epochs using Adam with a learning rate of $10^{-3}$ and cross-entropy loss. Averaged over five random seeds, the model achieved a test accuracy of $0.9786 \pm 0.0014$.

\raggedbottom

 After contraction with the first MPO core, the intermediate tensor has shape $(B,i_2,o_1,\chi)$, where $B$ denotes the batch size, $i_2$ and $o_1$ are the input dimension of the second MPO core and the output dimension of the first MPO core, respectively, and $\chi$ is the bond dimension. To obtain a single bond-space activation vector for each sample, we averaged over the non-bond dimensions, resulting in a tensor of shape $(B,\chi)$. For each bond feature $b$, we considered both tails of its activation distribution: the $k$ samples with the largest activations (positive tail) and the $k$ samples with the smallest activations (negative tail), with $k=100$. We computed the class purity independently for each tail and defined the best-tail purity of a bond feature as the maximum of the two, this approach keeps the signs separate since positive and negative activation can carry different class information:

$$ \mathrm{Purity}^{\pm}(b) = \frac{1}{k}\max_c N^{\pm}_{b,c}, $$ $$ \mathrm{BestPurity}(b) = \max\left\{ \mathrm{Purity}^{+}(b), \mathrm{Purity}^{-}(b) \right\}, $$

where $N^{+}{b,c}$ and $N^{-}{b,c}$ denote the number of samples from class $c$ among the $k$ largest and $k$ smallest activations of bond feature $b$, respectively. The dominant class of a bond feature is defined as the class attaining the maximum count in the tail yielding the highest purity. This formulation captures class selectivity expressed through either strongly positive or strongly negative bond activations.

Several bond features exhibited strong class selectivity, where the average best purity for all bond features is $0.8324 \pm 0.0440$. Table \ref{tab:bond_selectivity} shows some of the highly selective bond features.

\begin{table}[H]
\centering
\caption{Most selective bond features ranked by best-tail purity. 
The best tail corresponds to the activation direction (positive or negative) 
with the highest class best-tail purity among the top-$k$ samples, with $k=100$.}
\label{tab:bond_selectivity}
\begin{tabular}{cccccc}
\hline
\textbf{Bond Feature} &
\textbf{Dominant Digit} &
\textbf{Purity} &
\textbf{Best Tail} &
\textbf{Positive Purity} &
\textbf{Negative Purity} \\
\hline
0  & 0 & 1.00 & Negative & 0.63 & 1.00 \\
2  & 0 & 1.00 & Negative & 0.57 & 1.00 \\
7  & 6 & 1.00 & Positive & 1.00 & 0.99 \\
9  & 6 & 1.00 & Negative & 0.80 & 1.00 \\
14 & 0 & 1.00 & Positive & 1.00 & 0.73 \\
23 & 0 & 1.00 & Negative & 0.36 & 1.00 \\
25 & 6 & 1.00 & Positive & 1.00 & 0.93 \\
26 & 2 & 1.00 & Positive & 1.00 & 0.59 \\
12 & 2 & 0.99 & Positive & 0.99 & 0.90 \\
16 & 6 & 0.99 & Negative & 0.70 & 0.99 \\
19 & 4 & 0.99 & Positive & 0.99 & 0.54 \\
\hline
\end{tabular}
\end{table}

\begin{figure}[H]
    \centering

    \begin{subfigure}{0.48\columnwidth}
        \centering
        \includegraphics[width=\linewidth]{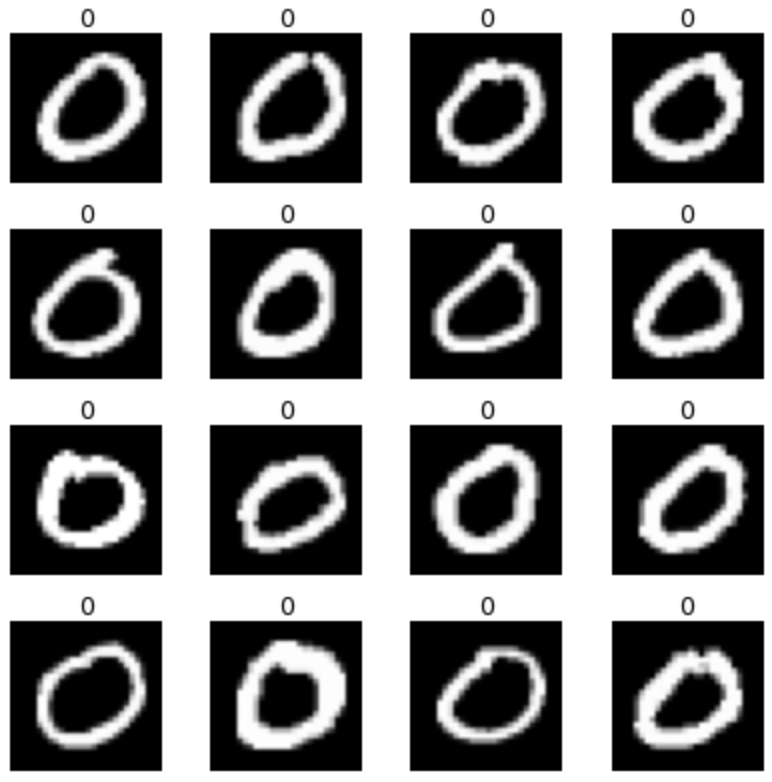}
        \caption{Bond feature 0}
        \label{fig:feature1}
    \end{subfigure}
    \hfill
    \begin{subfigure}{0.48\columnwidth}
        \centering
        \includegraphics[width=\linewidth]{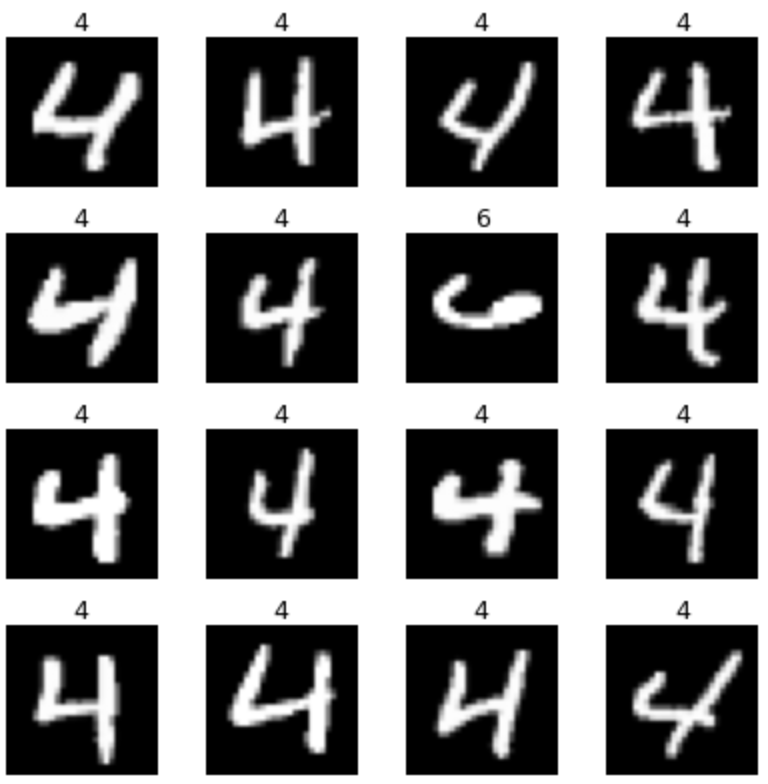}
        \caption{Bond feature 19}
        \label{fig:feature2}
    \end{subfigure}

    \vspace{0.3cm}

    \begin{subfigure}{0.48\columnwidth}
        \centering
        \includegraphics[width=\linewidth]{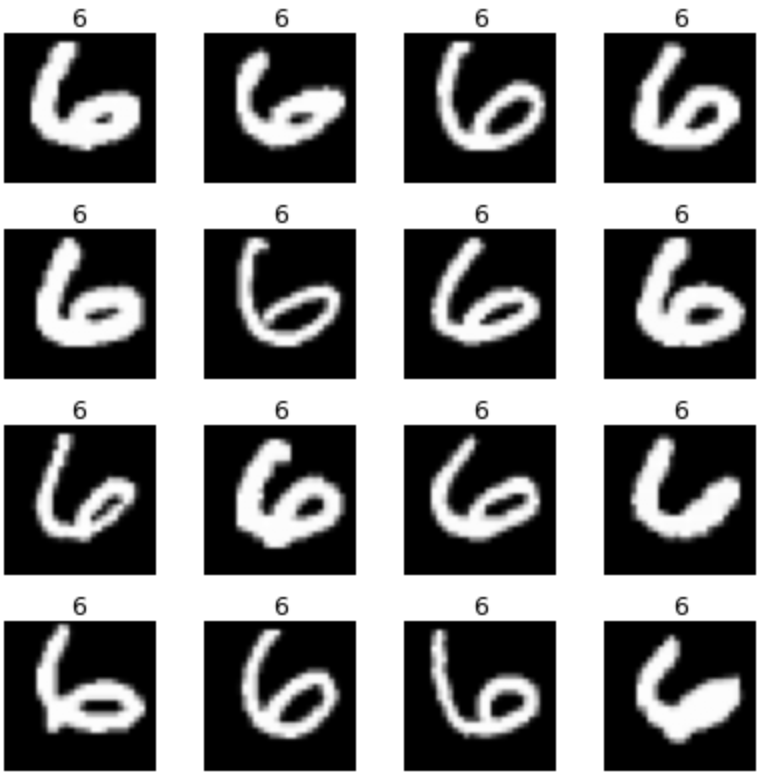}
        \caption{Bond feature 7}
        \label{fig:feature3}
    \end{subfigure}
    \hfill
    \begin{subfigure}{0.48\columnwidth}
        \centering
        \includegraphics[width=\linewidth]{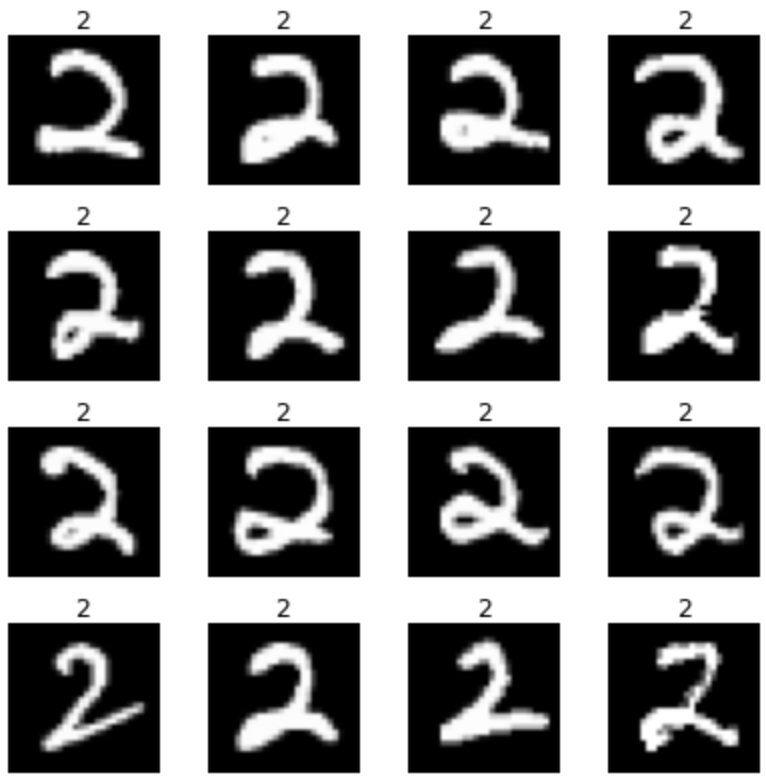}
        \caption{Bond feature 26}
        \label{fig:feature4}
    \end{subfigure}

    \caption{Top 16 activating test samples for four bond features from one of the five trained models. The average best-tail purity for this model is 0.7978.
}
    \label{fig:bond_features}
\end{figure}

These observations are consistent with bond features encoding class-relevant intermediate representations. The strong tail-selectivity observed in several bond dimensions suggests that task-relevant structure can be localized within the bond space. The stack representation makes these internal features directly accessible for analysis, offering a more detailed view of computation within tensorized layers than is typically available for dense architectures.

\begin{figure}[H]
    \centering
    \includegraphics[width=0.7\columnwidth]{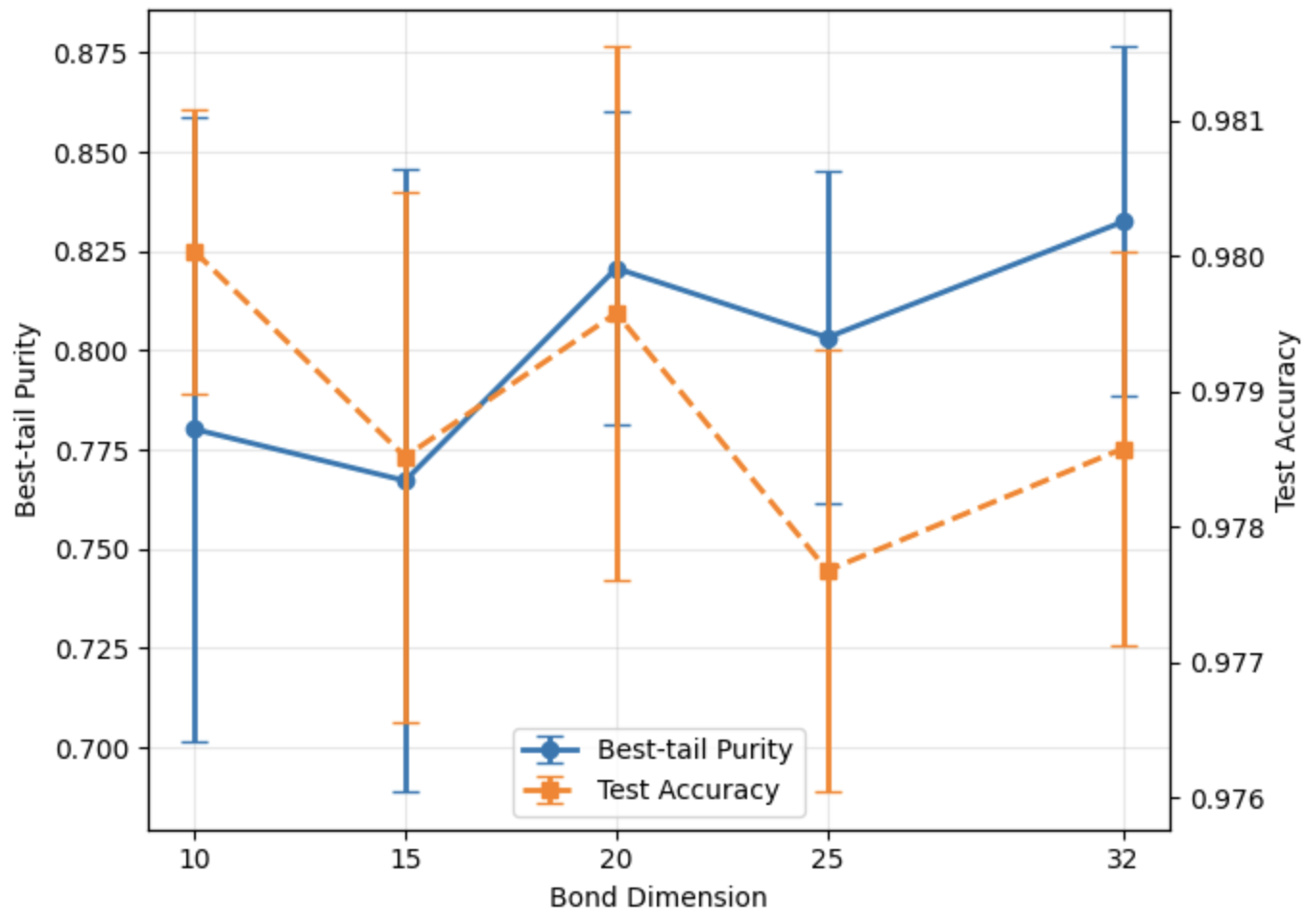}
    \caption{Purity and Test Accuracy vs Bond Dimension. top-k = 100}
    \label{fig:mnist-pur}
\end{figure}

We also investigate the effect of bond dimension on the class selectivity of features and how it correlates with test accuracy. Figure \ref{fig:mnist-pur} shows the best-tail bond-feature purity and test accuracy as a function of bond dimension, averaged over five random seeds. In this experiment, we set the minimum bond dimension to 10, since this equals the number of digits in the MNIST class. Test accuracy remains nearly constant across all bond dimensions, staying close to 0.98. Best-tail purity follows a non-monotonic pattern—dipping at bond dimension 15 before rising to its highest average value at bond dimension 32—though the error bars overlap substantially across all conditions, so this pattern should be interpreted with caution given the limited number of seeds (n=5). Overall, the results do not show a clear or consistent relationship between bond dimension and class selectivity in this setting.

\begin{figure}[H]
    \centering
    \includegraphics[width=0.7\columnwidth]{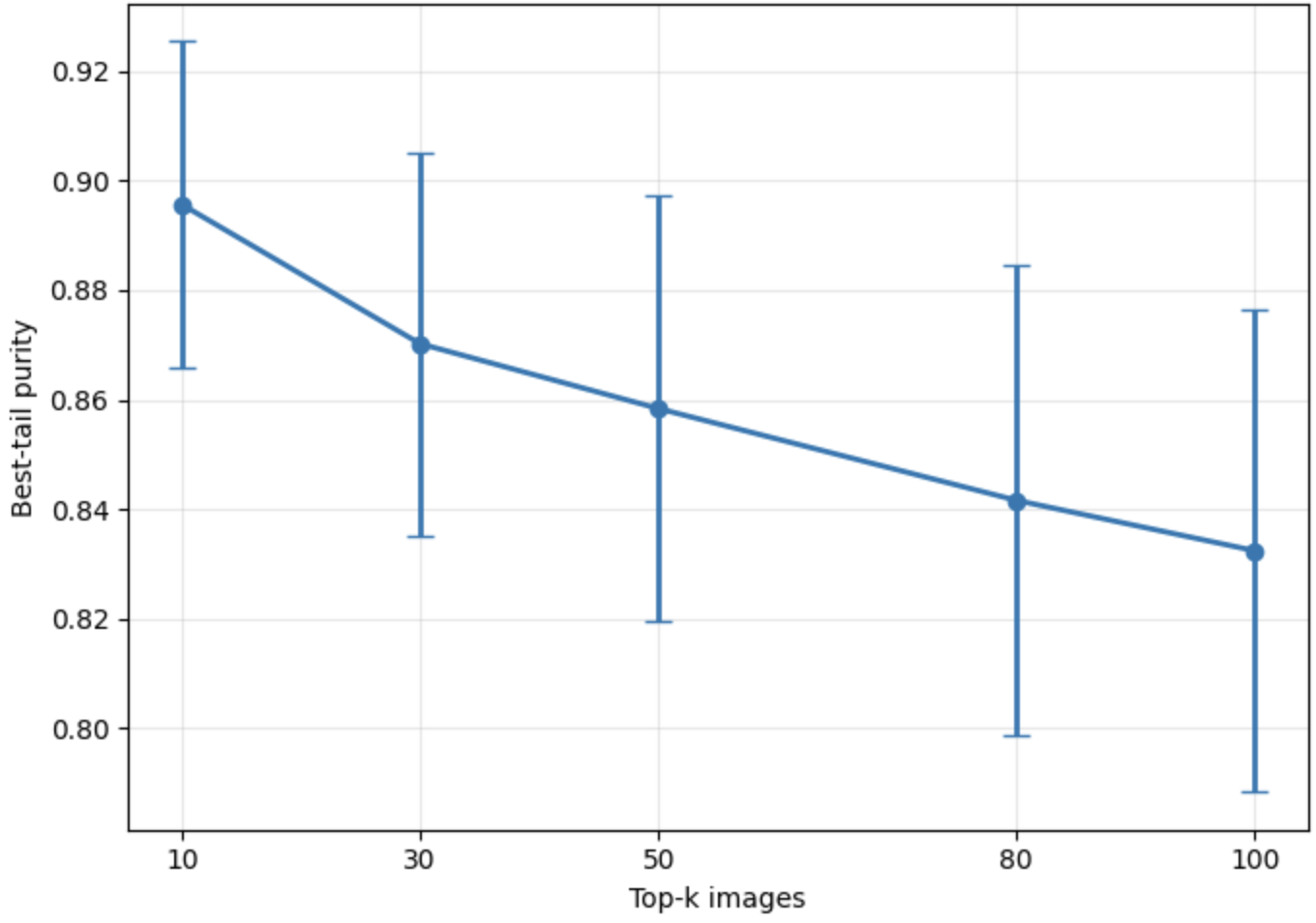}
    \caption{Best-tail purity vs top-k images. Bond dimension = 32}
    \label{fig:mnist-pur-topk}
\end{figure}

Figure \ref{fig:mnist-pur-topk} shows the effect of the number of selected top-k images on bond feature purity. We observe a negative correlation between best-tail purity and k, with the highest value achieved at k = 10. This is consistent with strongly activating samples being concentrated within the dominant class near the extremes of the activation distribution, though we note that purity computed over smaller top-k windows is expected to appear higher simply due to reduced sample size, independent of any class structure. As in the previous experiment, purity is averaged over five random seeds for a fairer comparison.

\textit{Experiment 2: Polysemanticity of Bond Features.}

To probe the interpretability of bond spaces, we use the same TNN architecture as in the class selectivity experiment, which includes a two-core MPO layer with bond dimension $\chi = 32$. We select the model that achieves a test accuracy of 0.9781 for one random seed. As before, after training we collect bond activations by contracting the input with the first MPO core and averaging over non-bond dimensions, yielding a $\chi$-dimensional activation vector per sample. We then compute the mean activation per class, producing a matrix $M \in \mathbb{R}^{10 \times 32}$, which we z-score normalize across classes for each column. A bond feature is said to \emph{fire} for a class if its absolute normalized activation exceeds $\tau = 1.0$, i.e., $|z| > 1$, corresponding to an activation that deviates by more than one standard deviation from the cross-class mean in either direction.

\paragraph{Results.}

The results in Table~\ref{tab:bond_superposition} indicate that, under this threshold, the MPO bond space represents class information predominantly through distributed, polysemantic features rather than dedicated class-specific detectors: none of the 32 bond features are monosemantic, and all are shared across at least two digit classes. Individual bond dimensions frequently encode information relevant to several digits at once, providing evidence of polysemanticity within the learned bond representation. The recurring overlap in the classes associated with different bond features (e.g., digits 0, 1, 2, and 8 co-occur across several features, including bonds 0, 11, and 24) further suggests that this polysemanticity is structured rather than arbitrary, potentially reflecting visual similarities—such as shared curves or strokes—between the corresponding digits. Confirming whether these shared features actually track such visual primitives would require further visualization of what each bond feature responds to at the pixel level, which we leave to future work.

\paragraph{Sensitivity to the selectivity threshold.}

\begin{figure}[t]
    \centering
        \includegraphics[width=0.7\columnwidth]{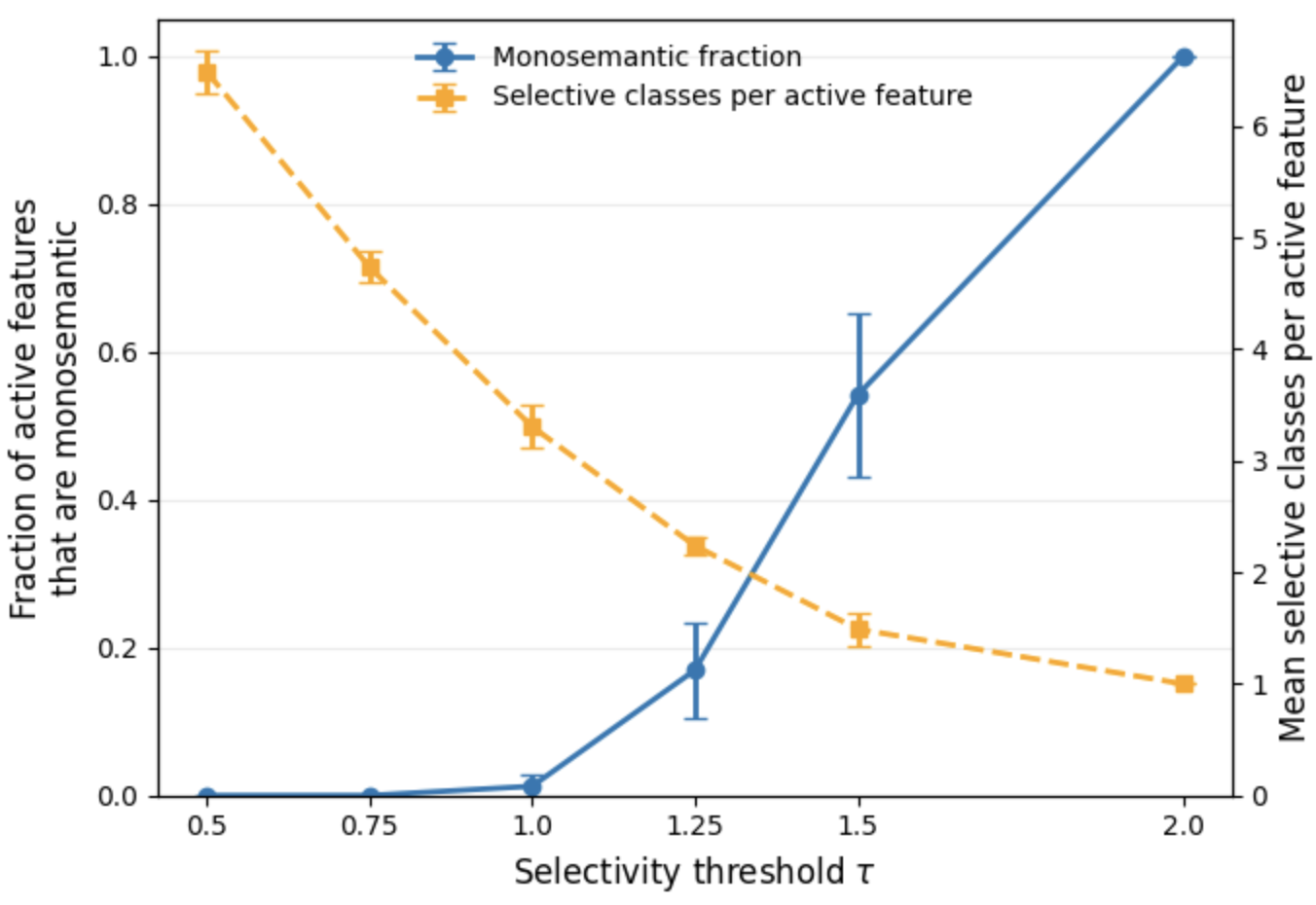}
        \caption{
        Effect of the selectivity threshold $\tau$ on bond-feature polysemanticity.
        The left axis shows the fraction of active bond features that are
        monosemantic, while the right axis shows the mean number of selective
        classes per active feature. Values are averaged over five random seeds,
        and error bars indicate one standard deviation.
        }
    \label{thresh}
\end{figure}

Because the polysemantic characterization above depends on a single fixed threshold, we repeated the analysis across $\tau \in \{0.5, 0.75, 1.0, 1.25, 1.5, 2.0\}$, averaged over five random seeds (Fig.~\ref{thresh}). At $\tau = 1.0$, our reported threshold, the bond space is almost entirely polysemantic, with $98.75\%$ of active features shared across multiple classes—consistent with the single-seed result in Table~\ref{tab:bond_superposition}. As $\tau$ increases, however, this picture shifts substantially: by $\tau = 1.5$, roughly half of active features are monosemantic, and by $\tau = 2.0$, every active feature is monosemantic in all five seeds (mono fraction $= 1.000 \pm 0.000$). This trend is mirrored by a steady decline in the mean number of selective classes per active feature, from $6.49$ at $\tau = 0.5$ to $1.00$ at $\tau = 2.0$. Notably, the number of features that remain active at all also shrinks substantially and somewhat inconsistently across seeds at high thresholds (ranging from 8 to 17 out of 32 at $\tau = 2.0$), so the perfect monosemantic fraction at this threshold reflects a smaller, seed-dependent subset of sharply-tuned features rather than the full bond space becoming class-specific.

Taken together, these results indicate that the polysemantic characterization of the bond space is threshold-dependent rather than a fixed property: at moderate activation levels, most features are shared across classes, while a smaller subset exhibits sharp, highly class-selective responses that only emerge as distinctly monosemantic under stricter thresholds. This is consistent with the tail-selectivity observed in the previous class selectivity experiment, where several bond features achieved near-perfect best-tail purity for a single class—suggesting that the same features responsible for that tail-selectivity may be the ones that "convert" to monosemantic status as $\tau$ increases. In short, the bond space is mostly polysemantic overall, but a small subset of features becomes sharply class-specific at extreme activations.

\begin{table}[t]
\centering
\caption{Bond feature polysemanticity at $\tau = 1.0$ for a single representative seed (see Fig.~\ref{thresh} for sensitivity to threshold choice). A bond feature is considered class-selective when $|z|>1$. Peak $|z|$ denotes the maximum absolute normalized activation across its selective classes.}
\label{tab:bond_superposition}
\begin{tabular}{c l c c}
\toprule
\textbf{Bond} & \textbf{Selective digits} & \textbf{Peak $|z|$} & \textbf{Type} \\
\midrule
0  & 0, 1, 2, 8, 9 & 1.93 & Shared \\
1  & 1, 2, 5, 6    & 2.00 & Shared \\
2  & 0, 1, 6       & 1.94 & Shared \\
3  & 1, 2, 9       & 1.88 & Shared \\
4  & 2, 5, 9       & 2.10 & Shared \\
5  & 1, 4, 5, 6, 8 & 2.10 & Shared \\
6  & 0, 1, 2, 9    & 1.90 & Shared \\
7  & 5, 6          & 2.24 & Shared \\
8  & 0, 1, 3, 4, 9 & 1.89 & Shared \\
9  & 3, 5, 6       & 1.93 & Shared \\
10 & 0, 3, 5, 7, 9 & 1.47 & Shared \\
11 & 0, 1, 2, 8    & 2.13 & Shared \\
12 & 0, 2, 4, 5    & 1.70 & Shared \\
13 & 0, 2, 6       & 1.98 & Shared \\
14 & 0, 1          & 2.48 & Shared \\
15 & 0, 1, 4       & 1.87 & Shared \\
16 & 6, 7, 9       & 2.15 & Shared \\
17 & 2, 6, 7       & 1.80 & Shared \\
18 & 0, 2, 6       & 2.00 & Shared \\
19 & 2, 4, 7       & 2.04 & Shared \\
20 & 0, 2          & 2.14 & Shared \\
21 & 0, 1, 8       & 1.71 & Shared \\
22 & 3, 5, 8       & 2.03 & Shared \\
23 & 0, 2, 8       & 2.24 & Shared \\
24 & 0, 1, 2, 8    & 1.98 & Shared \\
25 & 6, 7          & 2.47 & Shared \\
26 & 1, 2, 6       & 2.08 & Shared \\
27 & 0, 2, 4       & 2.35 & Shared \\
28 & 5, 8, 9       & 1.82 & Shared \\
29 & 1, 6, 7       & 1.88 & Shared \\
30 & 5, 7, 9       & 1.88 & Shared \\
31 & 2, 4, 6       & 2.28 & Shared \\
\midrule
\multicolumn{4}{c}{\textit{0 monosemantic, 32 shared (out of 32 total)}} \\
\bottomrule
\end{tabular}
\end{table}

\paragraph{Dense control baseline.} The class-selectivity and polysemanticity results reported above establish that bond features carry class-relevant structure, but do not by themselves indicate whether this structure is distinctive to the MPO's tensorized architecture or would emerge from any sufficiently trained intermediate representation of matched width. To disentangle these possibilities, we constructed a dense control network that replaces the two-core MPO layer with two ordinary linear layers of matching shapes: a $128 \to 32$ layer followed by a $32 \to 16$ layer, with no nonlinearity between them, mirroring the linear, structured-sparsity view of the MPO given by the stack representation (Sec.~3(2)). The resulting $32$-dimensional intermediate activation plays the same role as the MPO's bond space and is analyzed with an identical pipeline: the same encoder, training procedure, five random seeds, best-tail purity metric ($k=100$), and polysemanticity analysis across $\tau \in \{0.5, 0.75, 1.0, 1.25, 1.5, 2.0\}$.

Table~\ref{tab:dense_control} reports the results. The dense control achieves comparable test accuracy to the MPO ($0.9793 \pm 0.0023$ vs.\ $0.9786 \pm 0.0014$) and, notably, a \emph{higher} average best-tail purity ($0.9263 \pm 0.0311$ vs.\ $0.8324 \pm 0.0440$). The full threshold sweep tells a consistent story: the monosemantic fraction and mean selective-classes-per-feature track each other closely between the MPO and the dense control at every $\tau$, with the two curves nearly overlapping across the full range. This indicates that the class-selectivity and polysemanticity patterns observed in the bond space are not a distinctive consequence of the MPO's tensorized structure; a dense layer of matched width develops comparable, if not slightly stronger, class-selective structure under the same training regime. Although the dense control exhibits higher class selectivity under the specific metric considered here, this does not negate the distinctive structural advantages that TNNs offer for interpretability. In particular, the stack representation exposes multiple structured intermediate bond spaces within a single tensorized layer, providing a finer-grained view of how representations evolve through its internal computation. Moreover, the same tensor network admits multiple mathematically equivalent decompositions, enabling complementary views of the same input–output transformation. These properties are intrinsic to the tensorized architecture and are not captured by class selectivity alone. The present comparison therefore refines, rather than weakens, our position: TNNs should not be regarded as inherently more class-selective than dense representations, but as architectures that expose additional structured objects and levels of analysis for mechanistic investigation. Determining whether these structural advantages translate into more informative, stable, or causally meaningful explanations in larger models, non-classification tasks, and richer notions of feature interpretability remains an important direction for future work.

\begin{table}[t]
\centering
\caption{Dense control comparison. Accuracy and average best-tail purity ($k=100$), and monosemantic fraction / mean selective classes per active feature across selectivity thresholds $\tau$, for the MPO bond space versus a dense layer of matched width (32). All values are averaged over five random seeds.}
\label{tab:dense_control}
\begin{tabular}{lcc}
\toprule
 & \textbf{MPO} & \textbf{Dense control} \\
\midrule
Test accuracy & $0.9786 \pm 0.0014$ & $0.9793 \pm 0.0023$ \\
Best-tail purity ($k=100$) & $0.8324 \pm 0.0440$ & $0.9263 \pm 0.0311$ \\
\midrule
\multicolumn{3}{c}{\textit{Mono.\ fraction / classes-per-active feature, by $\tau$}} \\
\midrule
$\tau=0.50$ & $0.00$ / $6.49$ & $0.00$ / $6.49$ \\
$\tau=0.75$ & $0.00$ / $4.74$ & $0.00$ / $5.12$ \\
$\tau=1.00$ & $0.01$ / $3.31$ & $0.00$ / $3.62$ \\
$\tau=1.25$ & $0.17$ / $2.24$ & $0.10$ / $2.31$ \\
$\tau=1.50$ & $0.54$ / $1.49$ & $0.55$ / $1.49$ \\
$\tau=2.00$ & $1.00$ / $1.00$ & $0.98$ / $1.02$ \\
\bottomrule
\end{tabular}
\end{table}

\textit{TNNs vs. autoencoders.} 

Autoencoders build a single latent space by training an
encoder--decoder pair with an explicit reconstruction objective, so their bottleneck
representation is optimized purely to compress and recover the input, independently of
any downstream task. Bond indices in a TNN, in contrast, are not the output of a
dedicated compression module: they are intrinsic to the tensor network decomposition of
the weights and emerge directly from the network's original training objective, be it
classification, generation, or otherwise. This has two consequences for
interpretability. First, a TNN offers many bond spaces, one per internal tensor,
distributed across every tensorized layer, rather than a single bottleneck placed at one
point in the architecture, giving a much finer-grained trace of how representations
evolve. Second, because the same tensor network can be reorganized into different but
equivalent stacks (Sec.~3(2)), the same bond features admit multiple, mathematically
related ``views,'' a flexibility that a fixed autoencoder latent space does not provide.
These properties are complementary rather than competing: as discussed below, tensorizing an autoencoder itself lets its bottleneck
inherit this same fine-grained, stack-based decomposition, combining the disentangling
strengths of sparse autoencoders with the multi-scale structure native to TNNs.

\textit{Tensorized sparse autoencoders could provide finer interpretability of dense neural networks.}

Sparse autoencoders (SAEs) \cite{ng2011sparse} are widely
used as interpretability tools for analyzing large neural networks, particularly large language models \cite{cunningham2023sparse, lan2024sparse}. SAEs have demonstrated the
ability to resolve \emph{polysemanticity}–the phenomenon where
individual neurons encode multiple distinct features–by disentangling these features within a high-dimensional latent
space, and \emph{universality} \cite{cunningham2023sparse}, namely,
the emergence of universal features in the SAE latent space
across different LLMs.

A simple sparse autoencoder (SAE) consists of an encoder and a decoder defined as $L_x \equiv \mathrm{ENC}(x) = \mathrm{ReLU}(Ex + b)$ and $\mathrm{DEC}(L_x) = D L_x + c$, where $E$ and $D$ are the dense weight matrices of the encoder and decoder, $b$ and $c$ are the respective bias vectors, and $x$ is the input, typically an intermediate activation of the large language model (LLM) under study. The SAE is trained to approximately reconstruct its input, $\mathrm{DEC}(L_x) \approx x$, subject to a sparsity constraint on the encoder output, often implemented via an $L_1$-norm penalty on the batch loss. When successful, the SAE represents $x$ as a linear combination of the rows of the decoder matrix $D$, referred to as \emph{feature vectors}, with coefficients given by $L_x$.
In practice, these feature vectors often correspond to high-
level language concepts (e.g., ‘blue’, ‘fake’, ‘Golden Gate
Bridge’), allowing a polysemantic intermediate activation
to be disentangled into a sparse combination of (ideally)
monosemantic features, which enables interpretable narratives about the LLM’s processing.

Tensorizing a SAE and analyzing it via its stack representation provides an even finer lens on the disentangling
process (Fig. \ref{fig:forward_tensorization} (ii)). Encoder and decoder tensorizations
serve complementary roles: a tensorized encoder offers a
‘temporal resolution’ of how features are selected, while a
tensorized decoder provides a detailed decomposition of
the feature vectors. For instance, consider a tensorized encoder with layers $E1$ and $E2$, where $E1$’s output dimension equals the SAE’s latent dimension. The encoder output can then be viewed as the end of the sequence $L1_x \rightarrow L2_x \rightarrow L_x = \mathrm{ReLU}(L2_x)$, providing a detailed view of how the sparse vector $L_x$ emerges. In Fig. \ref{fig:forward_tensorization}(ii), the tensorized decoder contains a tree tensor network decomposition of the feature vectors, enabling comparisons of information-theoretic properties among semantically adjacent vectors. Analogous to quantum many-body systems, where tree tensor networks have been applied to resolve the wavefunction into length scales, top tensors (e.g., $D1$) inside a tree layer could capture global properties, while bottom tensors (e.g., $D3$) capture local distinctions. For
example, feature vectors corresponding to ‘red’ and ‘blue’
may share a top tensor encoding ‘color,’ with differences
reflected in the bottom tensors. This structure can also be
used during training to enforce hierarchical feature decomposition, e.g., by first training the decoder on ‘red’ instances,
freezing the top tensors, and continuing training with other
colors to separate features by scale.  

\textit{Information-theoretic interpretability.}

Beyond using (tensorized) autoencoders to probe neural network activations, tensor networks also offer a complementary, more information-theoretic perspective on interpretability. An accurate TN decomposition of a weight matrix, input
data, or activations produced inside a neural network, provides insight into how correlations are organized in these
tensors. TN representations naturally lend themselves to
precise quantification of correlation patterns–e.g., as the
von Neumann entropy \cite{aizpurua2025tensor}–and other
information-theoretic properties. Such properties could also
be leveraged to rank or categorize the training data itself. A
potentially fruitful direction for future research is to investigate how tensor network properties of weights, activations,
and data correlate with overall model performance.

\textbf{(8) Positioning TNNs Among Parameter-Efficient, Interpretable, and Efficient Architectures.\\}
\\Tensorized neural networks (TNNs) occupy a distinct position at the intersection of parameter-efficient modeling, architectural efficiency, and interpretability. Unlike structured parameterizations (e.g., low-rank factorization or pruning), which impose a fixed constraint on a dense layer, TNNs define a richer family of models through diverse tensor network geometries (e.g., tensor train, tree, ring). Through the stack representation, this structure is made explicit as a sequence of sparse linear transformations, rather than being implicitly encoded in a single matrix, hence enabling a finer analysis of features. Unlike efficient architectures such as convolutional or attention-based models—which rely on specific inductive biases—TNNs offer a more flexible design thanks to their large hyperparameter space, where different geometries can capture different dependencies (e.g, local or hierarchical). This means that TNNs do not impose a single inductive bias, but their architectural choice can be adapted based on the data type, enabling more flexibility. However, we note that the relationship between tensor network geometry and data remains an open question. At the same time, unlike post hoc interpretability methods, TNNs introduce interpretable structure directly in the model: The bond dimensions define intermediate feature spaces within each layer, enabling analysis of internal representations without additional tools.

\textbf{(9) Future outlook: Ultra-scalable TNNs via activation
function tensorization.}

While tensorization typically targets dense weight matrices, other components—nonlinear layers (e.g., activations,
normalization) and the activations themselves—can also be
tensorized. This is especially promising for large networks,
where intermediate activations often dominate memory and
compute costs. If activations remain tensorized throughout the network, training can proceed entirely in the TN
domain–inputs are tensorized, and each layer outputs tensorized activations, avoiding dense intermediates. Fig. \ref{fig:forward_tensorization} (iii)
illustrates such a forward pass. For example, the TN layer
receiving the MPO input $\{x_i\}$ can perform direct contraction without converting the tensor networks to dense tensors.
Such contractions may, however, grow the bond dimensions,
which must be truncated using TN approximation methods \cite{orus2019tensor, cirac2021matrix, camano2026successive}
 to
retain efficiency.

Special attention must be paid to the non-linear layers. Standard nonlinearities (e.g., ReLU, Tanh) operate pointwise
on dense data and are not directly compatible with TNs.
Applying them requires contracting the tensor network (e.g., $\{y_i\}$) in Fig. \ref{fig:forward_tensorization}(iii)) into a dense tensor, which is inefficient.
A better strategy is to design \emph{local} activation functions that
operate on individual tensors or small subsets within the network \footnote{However, naive local applications of standard activations can lead to unstable gradients and harm
training-for instance, local ReLU may zero out useful negative components prematurely.}.  Local activations must also be paired with suitable
normalization layers. An alternative is to use trainable local
nonlinearities, as in Kolmogorov-Arnold Networks (KANs)
\cite{liu2024kan}, which could jointly perform activation
and normalization while adapting to the structure of the
TN. A global activation function can be approximately applied to an MPO using cross-approximation, which avoids
explicit contraction of the tensor network \cite{oseledets2010tt, nunez2025learning}. However, this
approach can be computationally expensive. Therefore, local activation functions might offer a more efficient way to
introduce nonlinearity into the model.

\section{Alternative view: TNNs are not fit for
large-scale deployment}\label{alternative}

Despite their potential, TNNs remain underused in state of 
the art models. Below, we outline several key reasons and
possible directions for overcoming them.

\textbf{(1) Limited hardware and software support.}

Modern deep learning tools and hardware are optimized for dense architectures like transformers and CNNs.
For instance, CUDA and Triton kernels are easy to optimize for matrix multiplication. TNN layers rely on different types of computations such as performing a sequence of tensor contractions and reshaping. Additionally, current GPU software is highly optimized for regular dense operations such as GEMMs and convolutions. In contrast, hardware adaptation of TNNs is still challenging in practice and tensor operations are not yet efficiently supported
on GPUs. While libraries like NVIDIA’s cuTensor have improved TNN operations, significant gains are still needed for
implementing TNNs at scale. In dense networks, hardware
accelerators (e.g., FPGAs for LSTMs \cite{guo2025fpga} and
CNNs \cite{basalama2023flexcnn}) have led to major performance boosts. Similar hardware and software investment is
needed for TNNs. Tailoring tensorization to hardware — for
example, using hardware-aware neural architecture search
(HW-NAS)  \cite{chitty2022neural} or FPGA-
based optimization techniques like ALS \cite{nekooei2022compression} — could also improve performance.

\textbf{(2) Unclear inductive bias in deep learning.}

When used for compressing dense models, TNNs must approximate weights that lack a low-rank structure by design.
If such structure doesn’t emerge during training, post-hoc
tensorization may fail. However, conversely, training TNNs
from scratch could better leverage their native biases, potentially improving efficiency or accuracy — but this remains
underexplored. Understanding how the inherent structure of
data and the nature of the learning task give rise to emergent
low-rank features in model weights is crucial for further
progress.

\textbf{(3) Large and complex hyperparameter space.}

TNNs introduce a wide hyperparameter space – network
architecture, number of tensors, and bond dimensions. Selecting appropriate configurations is challenging and often
done heuristically, as systematic exploration is costly. We
need a better theoretical understanding of how performance
depends on these choices.

\textbf{(4) Integration with other compression methods.}

TNNs work best when combined with other techniques
like quantization. However, integrating them remains difficult. Most quantization methods target dense weights and
don’t generalize well to tensorized formats. Conversely,
tensorizing already-quantized weights restricts expressiveness. Solving this requires developing quantization schemes
tailored to tensor networks.

\textbf{(5) Open Questions.}

To guide future work, we identify a few key open questions:

1) \textit{Is the TNN inductive bias generally useful or limited to
specific tasks?}

Empirical studies applying tensorization across a variety of
model architectures suggest that TNNs provide a broadly
useful inductive bias \cite{wang2023tensor}. On the theoretical
side, some progress has been made in characterizing the
biases introduced by tensor network layers; see the recent
review \cite{borsoi2026low}. There are also concrete connections between certain tensor networks and well-studied
classes of neural networks. For example, \cite{ali2023approximation} demonstrates that tree tensor networks possess universal expressivity comparable to deep rectified linear unit
(ReLU) networks. Moreover, polynomial neural networks
can be represented exactly by tree tensor networks constructed from repeated copies of weights acting on copies
of the input, as illustrated in Fig. \ref{fig:tn_examples} (vii). To the best of our
knowledge, this specific correspondence has not been explicitly highlighted in the existing literature. We believe that
uncovering and formalizing such connections could provide
a principled foundation for designing TNN architectures
tailored to particular tasks.

2) \textit{Is it possible to infer optimal tensor decompositions directly from the structure of the data and task? Do real-world
modalities (e.g., text, images, audio) support efficient tensor
decompositions?}

As noted in the introduction, analyzing correlations in data
can help guide the design of more tailored TNNs, since the
geometry of a tensor network is directly tied to the correlation structure it can capture. Real-world data typically
exhibits intrinsic correlations, and it is reasonable to expect that these correlations should also be reflected in the
activations of a trained neural network. Consequently, the
weights must inherit a structure that preserves these correlations. For example, \cite{chen20252} studies correlations
in language datasets. According to the Hilberg conjecture,
language exhibits power-law correlations \cite{dkebowski2015relaxed, lin2016criticality}. \cite{cirac2021matrix} further shows
that the power-law exponent is close to 1, suggesting that
tensor trains are not well-suited for representing language
data. By contrast, images display area-law correlations,
which can be efficiently captured using PEPS. This highlights how different data modalities may naturally align with
different TNN geometries.

3) \textit{How should tensorization be combined with quantization
or pruning? For instance, what is the best way to quantize
MPO layers? How can quantized dense layers be tensorized
without major performance loss?}

There has been little systematic work on these questions.
Using the stack representation, one could, for example,
quantize the tensor factors individually by treating them
as separate linear layers, or alternatively, quantize a pretrained dense model and then tensorize it using algebraic
tools such as the Smith Normal Form \cite{stanley2016smith}. Similar considerations arise when combining pruning with ten-
sorization. What is the best strategy for pruning a tensorized
layer—pruning each tensor factor separately, or first contracting tensors, pruning the result, and then decomposing
back? More broadly, when compressing models, should
one tensorize first and then prune, or prune first and then
tensorize, to achieve faster and more accurate compression?

\textit{4) How can we optimize TNN execution for modern hardware?}

Hardware-aware strategies, including specialized chips designed for tensor contractions, show promise for accelerating TNN execution \cite{qu2021hardware, gu2022heat, taskynov2021tensor, liu2021efficient}. However, their
potential remains largely unexplored in the context of
production-scale models.

\begin{figure*}[t]
\centering

\resizebox{\textwidth}{!}{%
\begin{tikzpicture}[
    font=\footnotesize,
    node distance=0.8cm and 1.0cm,
    box/.style={
        rectangle,
        draw,
        rounded corners,
        align=left,
        text width=4.2cm,
        inner sep=5pt
    },
    arrow/.style={
        ->,
        thick
    }
]

\node[box] (established) {
\normalsize\textbf{Established}

\vspace{1mm}

\begin{itemize}
\setlength{\itemsep}{1pt}
\item Parameter efficiency (e.g., \cite{xu2023tensorgpt})
\item Inference speed-up (e.g., \cite{song2020speeding})
\item Stack representation for interpretability
      (Fig.~\ref{fig:stack_representation})
\end{itemize}
};

\node[box, right=of established] (promising) {
\normalsize\textbf{Promising, \\ less established}

\vspace{1mm}

\begin{itemize}
\setlength{\itemsep}{1pt}
\item Tensorized autoencoders for interpretability
\item Inductive bias from tensor structure
\item Hyperparameter design
      (bond dimension, tensor order, geometry)
\item Integration with other compression methods
\end{itemize}
};

\node[box, right=of promising] (future) {
\normalsize\textbf{Long-Term}

\vspace{1mm}

\begin{itemize}
\setlength{\itemsep}{1pt}
\item Tensorized nonlinearities
\item Hardware-aware design
\end{itemize}
};

\draw[arrow] (established.east) -- (promising.west);
\draw[arrow] (promising.east) -- (future.west);

\end{tikzpicture}%
}

\caption{
Maturity spectrum of tensorized neural networks (TNNs), from established foundations to emerging directions and long-term vision.
}
\label{fig:tnn_maturity}

\end{figure*}
\paragraph{Conclusions.} \

We argued that tensorization is a promising research direction for developing more scalable and theoretically grounded neural-network compression methods, while also providing additional structured representations that can be analyzed for interpretability. Our supporting experiments were conducted at the MNIST scale; while they illustrate the mechanisms we describe—compression–expressivity trade-offs, bond-feature selectivity, and polysemanticity—validating these observations on larger models and more complex data modalities (e.g., language or vision transformers) remains an important direction for future work. Given the current limitations and open questions, tensorization may have considerably more to offer if its properties are better understood and its practical challenges are systematically addressed. We therefore invite the research community to engage more deeply with this area and explore its potential to contribute to more accessible, interpretable, and computationally efficient artificial intelligence.


\bibliography{tmlr}
\bibliographystyle{tmlr}


\end{document}

%% file: math_commands.tex
\usepackage{amsmath,amsfonts,bm}

\def\eqref#1{equation~\ref{#1}}

\def\1{\bm{1}}

\DeclareMathAlphabet{\mathsfit}{\encodingdefault}{\sfdefault}{m}{sl}
\SetMathAlphabet{\mathsfit}{bold}{\encodingdefault}{\sfdefault}{bx}{n}

